\documentclass{article}

\usepackage{iclr2020_conference,times} %

\usepackage{amsmath,amsfonts,bm}

\def\eqref#1{equation~\ref{#1}}

\def\1{\bm{1}}

\DeclareMathAlphabet{\mathsfit}{\encodingdefault}{\sfdefault}{m}{sl}
\SetMathAlphabet{\mathsfit}{bold}{\encodingdefault}{\sfdefault}{bx}{n}

\usepackage{hyperref}
\usepackage{url}

\usepackage{amsmath,amssymb}
\usepackage{graphicx}
\usepackage{tabularx}
\usepackage{mlmacros}
\usepackage{todonotes}
\usepackage{caption}
\usepackage{subcaption}
\usepackage{cleveref}
\usepackage{tabularx}
\usepackage{booktabs}

\usepackage[T1]{fontenc}
\usepackage[acronym,nomain,
    style=list,
    nonumberlist,
    shortcuts,
    smallcaps,
]{glossaries}

\glsdisablehyper %

\newacronym{VAE}{vae}{variational auto-encoder}
\newacronym{SSM}{ssm}{state-space model}
\newacronym{SGD}{sgd}{stochastic gradient descent}
\newacronym{ELBO}{elbo}{evidence lower bound}
\newacronym{KL}{kl}{Kullback-Leibler}
\newacronym{LSTM}{lstm}{long short-term memory}
\newacronym{CNN}{cnn}{convolutional neural network}
\newacronym{RL}{rl}{reinforcement learning}

\newacronym{GECO}{geco}{Generalised \textsc{elbo} with Constrained Optimisation}

\newacronym{ADAM}{adam}{adam}
\glsunset{ADAM}

\newacronym[firstplural=multilayer perceptrons, plural=MLPs]{MLP}{mlp}{multilayer perceptron}
\newacronym{MC}{mc}{Monte Carlo}
\newacronym[firstplural=recurrent neural networks, plural=RNNs]{RNN}{rnn}{recurrent neural network}

\newacronym{MNIST}{mnist}{mnist}
\glsunset{MNIST}

\newacronym[firstplural=degrees of freedom, plural=DoFs]{DOF}{dof}{degree of freedom}

\newacronym{GENESIS}{genesis}{\textsc{gene}rative \textsc{s}cene \textsc{i}nference and \textsc{s}ampling}

\newacronym{GMM}{gmm}{Gaussian mixture model}
\newacronym{SBP}{sbp}{stick-breaking process}
\newacronym{IOU}{iou}{intersection over union}

\variables{u,x,z}
\probdists{p,q}

\title{Genesis: Generative Scene Inference and\\ Sampling with Object-Centric Latent\\ Representations}

\author{
Martin Engelcke
\!\thanks{Corresponding author: \texttt{martin@robots.ox.ac.uk}}
\ $^{\nabla}$, Adam R.~Kosiorek$^{\nabla \Delta}$, Oiwi Parker Jones$^{\nabla}$ \& Ingmar Posner$^{\nabla}$\\
$^{\nabla}$ Applied AI Lab, University of Oxford; 
$^{\Delta}$ Dept. of Statistics, University of Oxford
}

\iclrfinalcopy %
\begin{document}
\maketitle

\begin{abstract}

Generative latent-variable models are emerging as promising tools in robotics and reinforcement learning. Yet, even though tasks in these domains typically involve distinct objects, most state-of-the-art generative models do not explicitly capture the compositional nature of visual scenes. Two recent exceptions, \textsc{mon}et and \textsc{iodine}, decompose scenes into objects in an unsupervised fashion. Their underlying generative processes, however, do not account for component interactions. Hence, neither of them allows for principled sampling of novel scenes. Here we present \textsc{genesis}, the first  object-centric generative model of rendered 3D scenes capable of both decomposing \emph{and} generating scenes by capturing relationships between scene components. \textsc{genesis} parameterises a spatial \textsc{gmm} over images which is decoded from a set of object-centric latent variables that are either inferred sequentially in an amortised fashion or sampled from an autoregressive prior. We train \textsc{genesis} on several publicly available datasets and evaluate its performance on scene generation, decomposition, and semi-supervised learning.
\end{abstract}

\section{Introduction}
\label{sec:introduction}

Task execution in robotics and \gls{RL} requires accurate perception of and reasoning about discrete elements in an environment.
While supervised methods can be used to identify pertinent objects, it is intractable to collect labels for every scenario and task.
Discovering structure in data---such as objects---and learning to represent data in a compact fashion without supervision are long-standing problems in machine learning \citep{Comon1992ica,Tishby2000ib}, often formulated as \emph{generative latent-variable modelling} \citep[e.g.][]{kingma2013auto,rezende2014stochastic}.
Such methods have been leveraged to increase sample efficiency in \gls{RL} \citep{gregor2019shaping} and other supervised tasks \citep{steenkiste2019disentangle}.
They also offer the ability to \emph{imagine} environments for training \citep{ha2018world}.
Given the compositional nature of visual scenes, separating latent representations into object-centric ones can facilitate fast and robust learning \citep{watters2019cobra}, while also being amenable to relational reasoning \citep{Santoro2017relnet}.
Interestingly, however, state-of-the-art methods for generating realistic images do not account for this discrete structure \citep{brock2018large,parmar2018image}.

As in the approach proposed in this work, human visual perception is not passive. Rather it involves a creative interplay between external stimulation and an active, internal {generative model} of the world \citep{Rao_1999, Friston_2005}.
That this is necessary can be seen from the physiology of the eye, where the small portion of the visual field that can produce sharp images (\emph{fovea centralis}) motivates the need for rapid eye movements (\emph{saccades}) to build up a crisp and holistic percept of a scene \citep{Wandell_1995}. 
In other words, what we perceive is largely a mental simulation of the external world. 
Meanwhile, work in computational neuroscience tells us that visual features \citep[see, e.g.,][]{Hubel_1968} can be inferred from the statistics of static images using {unsupervised learning} \citep{Olshausen_1996}. 
Experimental investigations further show that specific brain areas (e.g.~LO) appear specialised for {objects}, for example responding more strongly to common objects than to scenes or textures, while responding only weakly to movement (cf.~MT) \citep[e.g.,][]{Grill_Spector_2004}.

In this work, we are interested in probabilistic generative models that can explain visual scenes compositionally via several latent variables.
This corresponds to fitting a probability distribution $\p{\bx}{}{\theta}$ with parameters $\theta$ to the data.
The compositional structure is captured by $K$ latent variables so that $\p{\bx}{}{\theta} = \int \p{\bx}{\bz_{1:K}}{\theta} \p{\bz_{1:K}}{}{\theta} \dint \bz_{1:K}$.
Models from this family can be optimised using the \gls{VAE} framework \citep{kingma2013auto,rezende2014stochastic}, by maximising a variational lower bound on the model evidence \citep{jordan1999introduction}.
\citet{burgess2019monet} and \citet{greff2019multi} recently proposed two such models, \textsc{mon}et and \textsc{iodine}, to decompose visual scenes into meaningful objects.
Both works leverage an \emph{analysis-by-synthesis} approach through the machinery of \glspl{VAE} \citep{kingma2013auto,rezende2014stochastic} to train these models without labelled supervision, e.g. in the form of ground truth segmentation masks.
However, the models have a factorised prior that treats scene components as independent.
Thus, neither provides an object-centric generation mechanism that accounts for relationships between constituent parts of a scene, e.g. two physical objects cannot occupy the same location, prohibiting the component-wise generation of novel scenes and restricting the utility of these approaches.
Moreover, \textsc{mon}et embeds a \gls{CNN} inside of an \gls{RNN} that is unrolled for each scene component, which does not scale well to more complex scenes.
Similarly, \textsc{iodine} utilises a \gls{CNN} within an expensive, gradient-based iterative refinement mechanism.

Therefore, we introduce \gls{GENESIS} which is, to the best of our knowledge, the first object-centric generative model of rendered 3D scenes capable of both decomposing and generating scenes\footnote{We use the terms ``object'' and ``scene component'' synonymously in this work.}.
Compared to previous work, this renders \gls{GENESIS} significantly more suitable for a wide range of applications in robotics and reinforcement learning. %
\gls{GENESIS} achieves this by modelling relationships between scene components with an expressive, autoregressive prior that is learned alongside a sequential, amortised inference network.
Importantly, sequential inference is performed in low-dimensional latent space, allowing all convolutional encoders and decoders to be run in parallel to fully exploit modern graphics processing hardware.

We conduct experiments on three canonical and publicly available datasets: \emph{coloured Multi-dSprites} \citep{burgess2019monet}, the \emph{GQN} dataset \citep{eslami2018neural}, and \emph{ShapeStacks} \citep{groth2018shapestacks}.
The latter two are simulated 3D environments which serve as testing grounds for navigation and object manipulation tasks, respectively.
We show both qualitatively and quantitatively that in contrast to prior art, \gls{GENESIS} is able to generate coherent scenes while also performing well on scene decomposition.
Furthermore, we use the scene annotations available for ShapeStacks to show the benefit of utilising general purpose, object-centric latent representations from \gls{GENESIS} for tasks such as predicting whether a block tower is stable or not.

Code and models are available at \url{https://github.com/applied-ai-lab/genesis}.

\section{Related Work}
\label{sec:related}

\textbf{Structured Models}\ \ 
Several methods leverage structured latent variables to discover objects in images without direct supervision.
\textsc{cst-vae} \citep{huang2015efficient}, \textsc{air} \citep{eslami2016attend}, \textsc{sqair} \citep{kosiorek2018sqair}, and \textsc{spair} \citep{crawford2019spatially} use spatial attention to partition scenes into objects.
\textsc{tagger} \citep{greff2016tagger}, \textsc{nem} \citep{greff2017neural}, and \textsc{r-nem} \citep{van2018relational} perform unsupervised segmentation by modelling images as spatial mixture models.
\textsc{scae} \citep{kosiorek2019stacked} discovers geometric relationships between objects and their parts by using an affine-aware decoder.
Yet, these approaches have not been shown to work on more complex images, for example visual scenes with 3D spatial structure, occlusion, perspective distortion, and multiple foreground and background components as considered in this work.
Moreover, none of them demonstrate the ability to generate novel scenes with relational structure.

While \citet{xu2018multi} present an extension of \citet{eslami2016attend} to generate images, their method only works on binary images with a uniform black background and assumes that object bounding boxes do not overlap.
In contrast, we train \gls{GENESIS} on rendered 3D scenes from \citet{eslami2018neural} and \citet{groth2018shapestacks} which feature complex backgrounds and considerable occlusion to perform both decomposition \emph{and} generation.
Lastly, \cite{xu2019unsupervised} use ground truth pixel-wise flow fields as a cue for segmenting objects or object parts.
Similarly, \gls{GENESIS} could be adapted to also leverage temporal information which is a promising avenue for future research.

\clearpage

\textbf{\textsc{mon}et \& \textsc{iodine}}\ \
While this work is most directly related to \textsc{mon}et \citep{burgess2019monet} and \mbox{\textsc{iodine}} \citep{greff2019multi}, it sets itself apart by introducing a generative model that captures relations between scene components with an autoregressive prior, enabling the unconditional generation of coherent, novel scenes.
Moreover, \textsc{mon}et relies on a deterministic attention mechanism rather than utilising a proper probabilistic inference procedure.
This implies that the training objective is not a valid lower bound on the marginal likelihood and that the model cannot perform density estimation without modification.
Furthermore, this attention mechanism embeds a \gls{CNN} in a \gls{RNN}, posing an issue in terms of scalability.
These two considerations do not apply to \textsc{iodine}, but \textsc{iodine} employs a gradient-based, iterative refinement mechanism which expensive both in terms of computation and memory, limiting its practicality and utility.
Architecturally, \gls{GENESIS} is more similar to \textsc{mon}et and does not require expensive iterative refinement as \textsc{iodine}.
Unlike \textsc{mon}et, though, the convolutional encoders and decoders in \gls{GENESIS} can be run in parallel, rendering the model computationally more scalable to inputs with a larger number of scene components.

\textbf{Adversarial Methods}\ \
A few recent works have proposed to use an adversary for scene segmentation and generation.
\citet{chen2019unsupervised} and \citet{bielski2019emergence} segment a single foreground object per image and \citet{arandjelovic2019object} segment several synthetic objects superimposed on natural images.
\citet{azadicompositional} combine two objects or an object and a background scene in a sensible fashion and \citet{van2018case} can generate scenes with a potentially arbitrary number of components.
In comparison, \gls{GENESIS} performs both inference and generation, does not exhibit the instabilities of adversarial training, and offers a probabilistic formulation which captures uncertainty, e.g. during scene decomposition.
Furthermore, the complexity of \gls{GENESIS} increases with $\mathcal{O}(K)$, where $K$ is the number of components, as opposed to the $\mathcal{O}(K^2)$ complexity of the \emph{relational stage} in \citet{van2018case}.

\textbf{Inverse Graphics}\ \ 
A range of works formulate scene understanding as an inverse graphics problem.
These well-engineered methods, however, rely on scene annotations for training and lack probabilistic formulations.
For example, \citet{wu2017neural} leverage a graphics renderer to decode a structured scene description which is inferred by a neural network.
\citet{romaszko2017vision} pursue a similar approach but instead make use of a differentiable graphics render.
\citet{wu2017learning} further employ different physics engines to predict the movement of billiard balls and block towers.

\section{\textsc{Genesis}: \textsc{Gene}rative \textsc{S}cene \textsc{I}nference and \textsc{S}ampling}
\label{sec:compositional}

\begin{figure}[t!]
    \centering
    \includegraphics[trim=0 0 0 0, clip, width=1.0\textwidth]{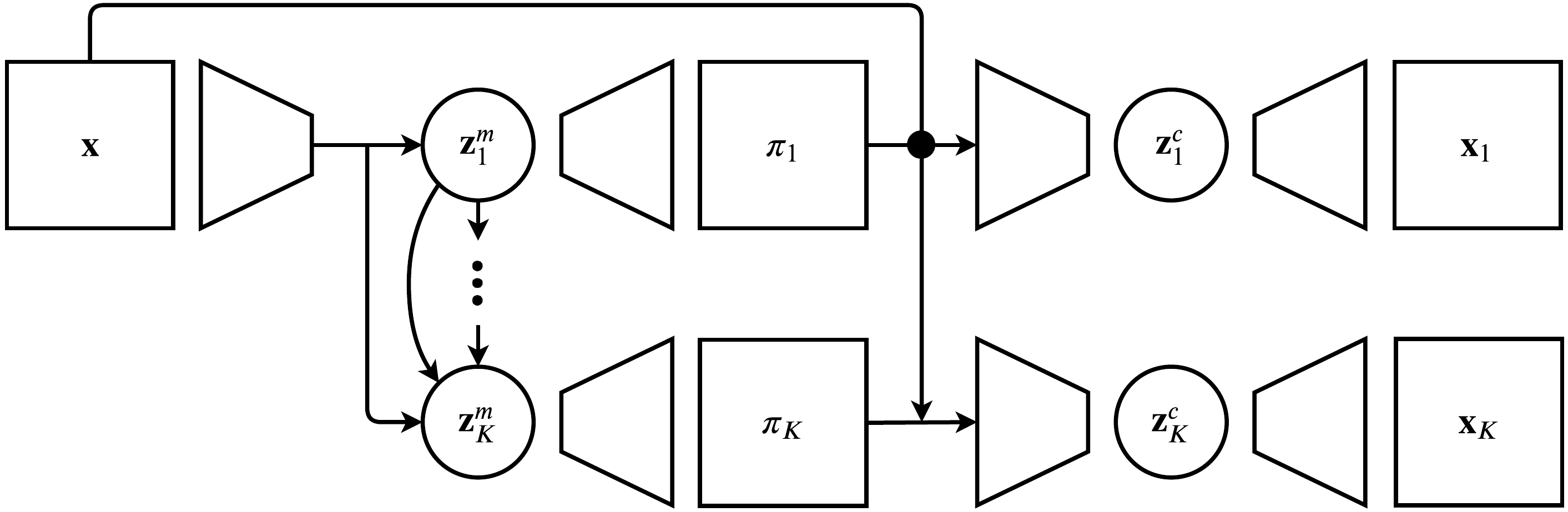}
    \caption{\gls{GENESIS} illustration. Given an image $\bx$, an encoder and an \gls{RNN} compute the mask latents $\bz^m_k$. These are decoded to obtain the mixing probabilities $\mathbf{\pi}_k$. The image and individual masks are concatenated to infer the component latents $\bz^c_k$ from which the scene components $\bx_k$ are decoded.
    }
    \label{fig:arch}
\end{figure}

\begin{figure}
	\centering
	\includegraphics[width=0.85\textwidth]{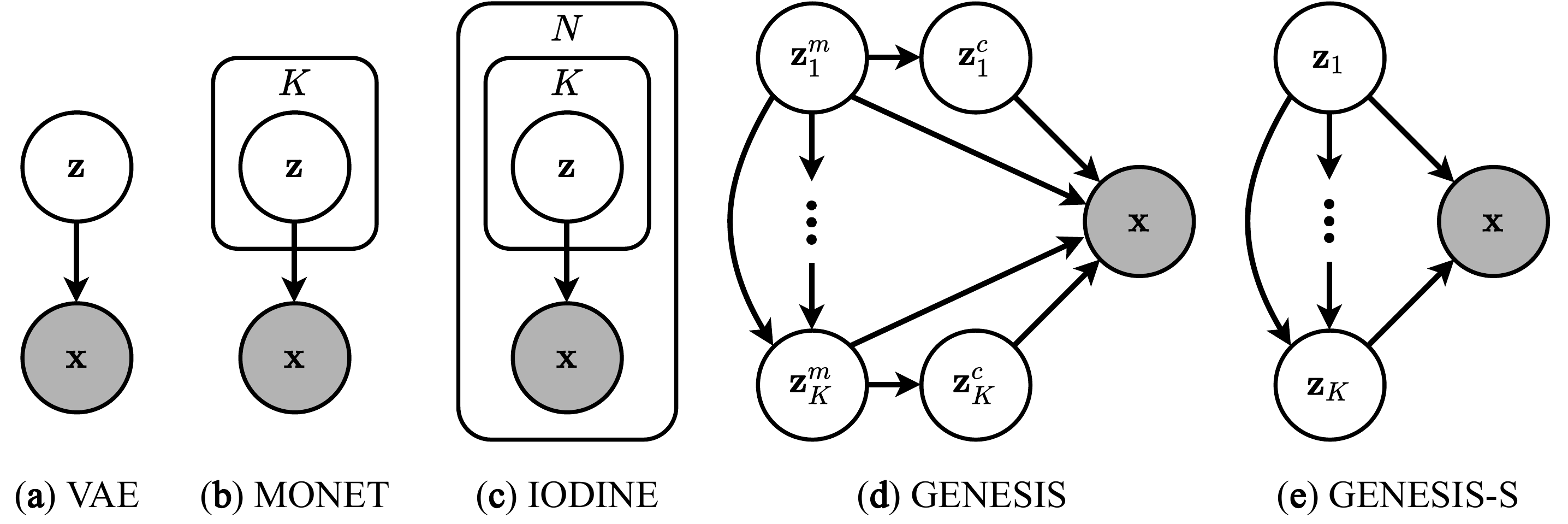}
    \caption{Graphical model of \textsc{genesis} compared to related methods. $N$ denotes the number of refinement iterations in \textsc{iodine}. Unlike the other methods, both \textsc{genesis} variants explicitly model dependencies between scene components.}
\label{fig:pgms}
\end{figure}

In this section, we first describe the generative model of \gls{GENESIS} and a simplified variant called \gls{GENESIS}\textsc{-s}.
This is followed by the associated inference procedures and two possible learning objectives.
\gls{GENESIS} is illustrated in \Cref{fig:arch} and \Cref{fig:pgms} shows the graphical model in comparison to alternative methods.
An illustration of \gls{GENESIS}\textsc{-s} is included \Cref{app:genesis_architecture}, \Cref{fig:arch_s}.

\paragraph{Generative model}
Let $\bx \in \RR^{H \times W \times C}$ be an image.
We formulate the problem of image generation as a spatial \gls{GMM}.
That is, every Gaussian component $k=1,\dots,K$ represents an image-sized scene component $\bx_k \in \RR^{H \times W \times C}$.
$K \in \mathbb{N}_+$ is the maximum number of scene components.
The corresponding \emph{mixing probabilities} $\pi_k \in [0, 1]^{H \times W}$ indicate whether the component is present at a location in the image.
The mixing probabilities are normalised across scene components, i.e. $\forall_{i,j} \sum_k \pi_{i,j,k} = 1$, and can be regarded as spatial \emph{attention masks}.
Since there are strong spatial dependencies between components, we formulate an autoregressive prior distribution over mask variables $\bz^m_k \in \RR^{D_m}$ which encode the mixing probabilities $\pi_k$, as
\begin{equation}
    \p{\bz_{1:K}^m}{}{\theta}
        = \prod_{k=1}^K \p{\bz^m_k}{\bz^m_{1:k-1}}{\theta}
        = \prod_{k=1}^K \p{\bz^m_k}{\bu_k}{\theta} \lvert_{\bu_k = \operatorname{R}_\theta (\bz^m_{k-1}, \bu_{k-1})}\,.
\end{equation}
The dependence on previous latents $\bz^m_{1:k-1}$ is implemented via an \gls{RNN} $\operatorname{R}_\theta$ with hidden state $\bu_k$.

Next, we assume that the scene components $\bx_k$ are conditionally independent given their spatial allocation in the scene.
The corresponding conditional distribution over component variables \mbox{$\bz^c_k \in \RR^{D_c}$} which encode the scene components $\bx_k$ factorises as follows,
\begin{equation}
    \p{\bz^c_{1:K}}{\bz^m_{1:K}}{\theta} = \prod_{k=1}^K \p{\bz^c_k}{\bz^m_k}{\theta}\,. \label{eq:conditional_appearance}
\end{equation}
Now, the image likelihood is given by a mixture model,
\begin{equation}
    \p{\bx}{\bz^m_{1:K}, \bz^c_{1:K}} = \sum_{k=1}^K \pi_k\, \p{\bx_k}{\bz^c_k}{\theta}\,, \label{eq:gauss_mixture}
\end{equation}
where the mixing probabilities $\pi_k = \operatorname{\pi_\theta} (\bz^m_{1:k})$ are created via a \glsreset{SBP}\gls{SBP} adapted from \citet{burgess2019monet} as follows, slightly overloading the $\pi$ notation,
\begin{equation}
    \pi_1 = \operatorname{\pi_\theta} (\bz^m_1)\,,
    \qquad
    \pi_k = \left(1 - \sum_{j=1}^{k-1} \pi_j \right) \operatorname{\pi_\theta} (\bz^m_k)\,,
    \qquad
    \pi_K = \left(1 - \sum_{j=1}^{K-1} \pi_j \right)
    \,. \label{eq:sbp}
\end{equation}
Note that this step is not necessary for our model and instead one could use a $\operatorname{softmax}$ to normalise masks as in \citet{greff2019multi}. 

Finally, omitting subscripts, the full generative model can be written as
\begin{equation}
    \p{\bx}{}{\theta} = \iint \p{\bx}{\bz^c,\bz^m}{\theta} \p{\bz^c}{\bz^m}{\theta} \p{\bz^m}{}{\theta} \dint \bz^m \dint \bz^c\,,
\end{equation}
where we assume that all conditional distributions are Gaussian.
The Gaussian components of the image likelihood have a fixed scalar standard deviation $\sigma^2_x$.
We refer to this model as \gls{GENESIS}.
To investigate whether separate latents for masks and component appearances are necessary for decomposition, we consider a simplified model, \gls{GENESIS}\textsc{-s}, with a single latent variable per component,
\begin{equation}
	\p{\bz_{1:K}}{}{\theta} = \prod_{k=1}^K \p{\bz_k}{\bz_{1:k-1}}{\theta}.
\end{equation}
In this case, $\bz_k$ takes the role of $\bz_k^c$ in \Cref{eq:gauss_mixture} and of $\bz_k^m$ in \Cref{eq:sbp}, while \Cref{eq:conditional_appearance} is no longer necessary.

\paragraph{Approximate posterior}
We amortise inference by using an approximate posterior distribution with parameters $\phi$ and a structure similar to the generative model. 
The full approximate posterior reads as follows,
\begin{equation}
	\begin{aligned}
	&\q{\bz_{1:K}^c, \bz_{1:K}^m}{\bx}{\phi} = \q{\bz_{1:K}^m}{\bx}{\phi}\, \q{\bz_{1:K}^c}{\bx, \bz_{1:K}^m}{\phi}\,,\quad \text{where}\\
	\q{\bz_{1:K}^m}{\bx}{\phi} = &\prod_{k=1}^K \q{\bz_k^m}{\bx, \bz_{1:k-1}^m}{\phi}\,,
	\quad \text{and} \quad
	\q{\bz_{1:K}^c}{\bx, \bz_{1:K}^m}{\phi} = \prod_{k=1}^K \q{\bz_k^c}{\bx, \bz_{1:k}^m}{\phi}\,,
	\end{aligned}
\end{equation}
with the dependence on $\bz^m_{1:k-1}$ realised by an \gls{RNN} $\operatorname{R_\phi}$. The \gls{RNN} could, in principle, be shared with the prior, but we have not investigated this option. All conditional distributions are Gaussian.
For \gls{GENESIS}\textsc{-s}, the approximate posterior takes the form $\q{\bz_{1:K}}{\bx}{\phi} = \prod_{k=1}^K \q{\bz_k}{\bx, \bz_{1:k-1}}{\phi}$\,.

\paragraph{Learning}
\Gls{GENESIS} can be trained by maximising the \glsreset{ELBO}\gls{ELBO} on the log-marginal likelihood $\log\p{\bx}{}{\theta}$, given by
\begin{align}
    \loss[\textsc{elbo}]{\bx} &= \expc{\log \frac{ \p{\bx}{\bz^c,\bz^m}{\theta} \p{\bz^c}{\bz^m}{\theta} \p{\bz^m}{}{\theta}}{\q{\bz^c}{\bz^m,\bx}{\phi} \q{\bz^m}{\bx}{\phi}} }{}{{\q{\bz^c,\bz^m}{\bx}{\phi}}} \label{eq:elbo}\\
    &= \expc{\log\p{\bx}{\bz^c,\bz^m}{\theta} }{}{{\q{\bz^c,\bz^m}{\bx}{\phi}}} -
    \operatorname{\textsc{kl}} \left(
        \q{\bz^c,\bz^m}{\bx}{\phi} \mid\mid \p{\bz^c,\bz^m}{}{\theta}
    \right) \label{eq:decomposed_elbo}
    \,.
\end{align}
However, this often leads to a strong emphasis on the likelihood term, while allowing the marginal approximate posterior $\q{\bz}{}{\phi} = \expc{ \q{\bz}{\bx}{\phi} }{}{\p{\bx}{}{\textrm{data}}}$ to drift away from the prior distribution, hence increasing the \textsc{kl}-divergence.
This also decreases the quality of samples drawn from the model.
To prevent this behaviour, we use the \gls{GECO} objective from \citet{rezende2018taming} instead, which changes the learning problem to minimising the \textsc{kl}-divergence subject to a reconstruction constraint.
Let $C \in \RR$ be the minimum allowed reconstruction log-likelihood, \gls{GECO} then uses Lagrange multipliers to solve the following problem,
\begin{equation}
    \begin{aligned}
    \theta^\star, \phi^\star &= \arg\min_{\theta, \phi}
    \operatorname{\textsc{kl}} \left(
        \q{\bz^c,\bz^m}{\bx}{\phi} \mid\mid \p{\bz^c,\bz^m}{}{\theta}
    \right)\\
    &\text{such that}\quad
    \expc{\log\p{\bx}{\bz^c,\bz^m}{\theta} }{}{{\q{\bz^c,\bz^m}{\bx}{\phi}}} \geq C\,.
    \end{aligned}
\end{equation}

\section{Experiments}
\label{sec:experimental_results}

In this section, we present qualitative and quantitative results on \emph{coloured Multi-dSprites} \citep{burgess2019monet}, the ``rooms-ring-camera'' dataset from \emph{GQN} \citep{eslami2018neural} and the \emph{ShapeStacks} dataset \citep{groth2018shapestacks}.
We use an image resolution of 64-by-64 for all experiments.
The number of components is set to $K=5$, $K=7$, and $K=9$ for Multi-dSprites, GQN, and \mbox{ShapeStacks}, respectively.
More details about the datasets are provided in \Cref{app:datasets}.
Implementation and training details of all models are described in \Cref{app:implementation}.

\subsection{Component-Wise Scene Generation}

Unlike previous works, \gls{GENESIS} has an autoregressive prior to capture intricate dependencies between scene components.
Modelling these relationships is necessary to generate coherent scenes.
For example, different parts of the background need to fit together; we do not want to create components such as the sky several times; and several physical objects cannot be in the same location.
\gls{GENESIS} is able to generate novel scenes by sequentially sampling scene components from the prior and conditioning each new component on those that have been generated during previous steps.

\begin{figure}[h!]
    \centering
    \includegraphics[width=1.0\textwidth]{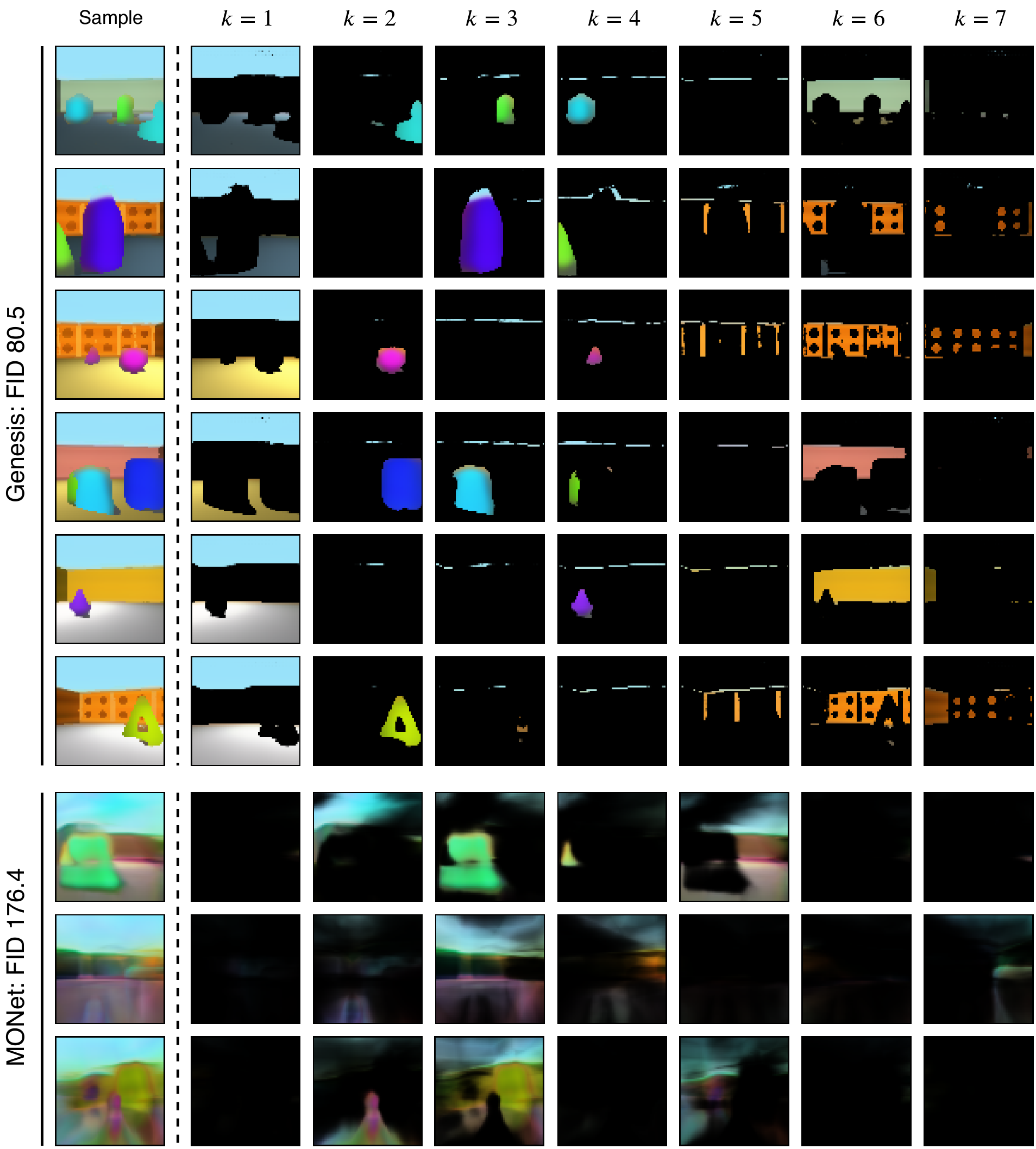}
    \caption{Component-by-component scene generation with \gls{GENESIS} and \textsc{mon}et after training on the GQN dataset. The first pane shows the final scene and the subsequent panes show the components generated at each step. \gls{GENESIS} first generates the sky and the floor, followed by individual objects, and finally distinct parts of the wall in the background to compose a coherent scene. \textsc{mon}et, in contrast, only generates incomplete components that do not fit together.}
    \label{fig:generation_gqn}
\end{figure}

After training \gls{GENESIS} and \textsc{mon}et on the GQN dataset, \Cref{fig:generation_gqn} shows the component-by-component generation process of novel scenes, corresponding to drawing samples from the respective prior distributions.
More examples of generated scenes are shown in \Cref{fig:generation_gqn_g32m16}, \Cref{app:generation}.
With \gls{GENESIS}, either an object in the foreground or a part of the background is generated at every step and these components fit together to make up a semantically consistent scene that looks similar to the training data.
\textsc{mon}et, though, generates random artefacts at every step that do not form a sensible scene. 
These results are striking but not surprising: \textsc{mon}et was not designed for scene generation. The need for such a model is why we developed \gls{GENESIS}.

\clearpage

Notably, \gls{GENESIS} pursues a consistent strategy for scene generation:
Step one generates the floor and the sky, defining the layout of the scene.
Steps two to four generate individual foreground objects.
Some of these slots remain empty if less than three objects are present in the scene.
The final three steps  generate the walls in the background.
We conjecture that this strategy evolves during training as the floor and sky constitute large and easy to model surfaces that have a strong impact on the reconstruction loss.
Finally, we observe that some slots contain artefacts of the sky at the top of the wall boundaries.
We conjecture this is due to the fact that the mask decoder does not have skip connections as typically used in segmentation networks, making it difficult for the model to predict sharp segmentation boundaries.
Scenes generated by \gls{GENESIS}\textsc{-s} are shown in \Cref{fig:generation_gqn_gs6} and \Cref{fig:generation_gqn_gs_4x8}, \Cref{app:generation}. While \gls{GENESIS}\textsc{-s} does separate the foreground objects from the background, it generates them in one step and the individual background components are not very interpretable.

\clearpage

\subsection{Inference of Scene Components}

\begin{figure}[h]
    \centering
    \includegraphics[width=1.0\textwidth]{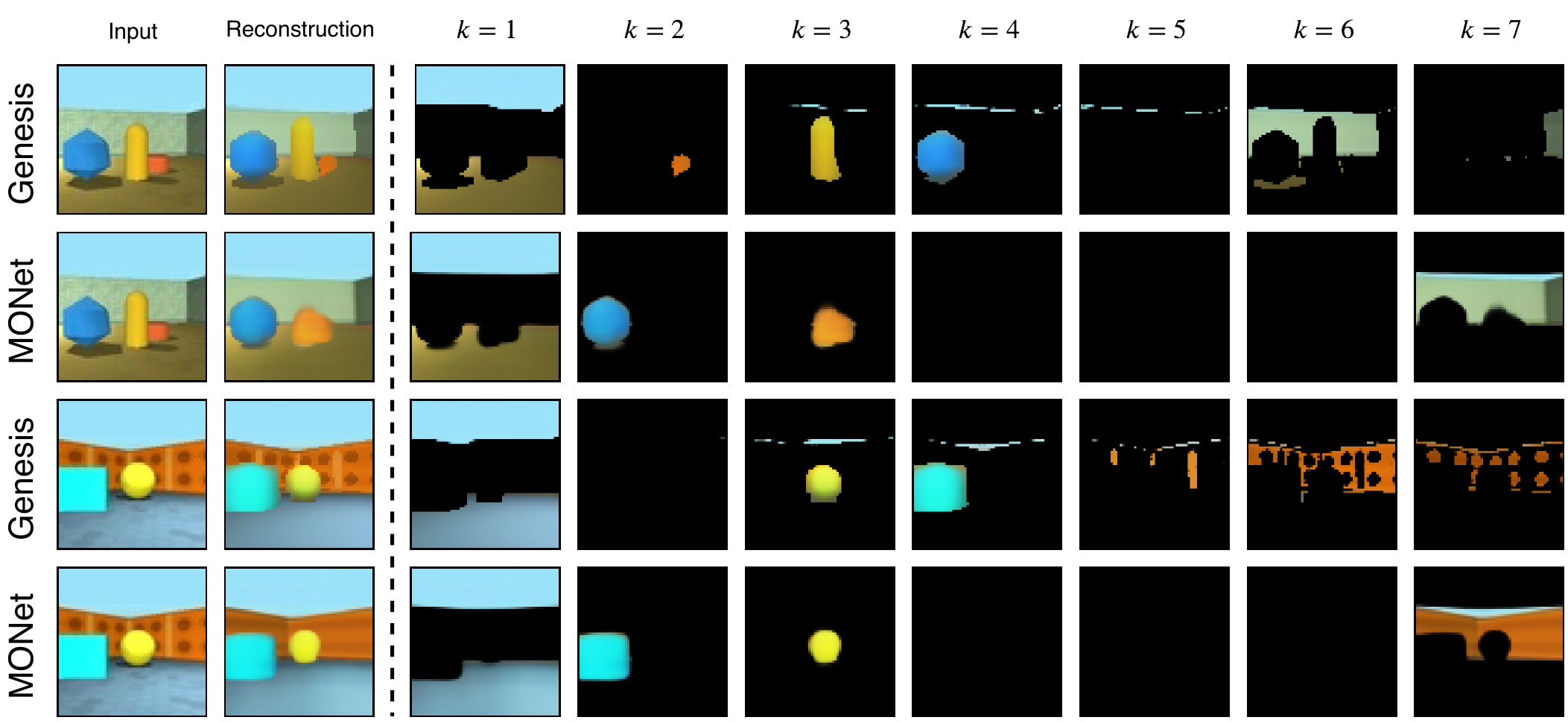}
    \caption{Step-by-step decomposition of the same scene from GQN with \gls{GENESIS} and \textsc{mon}et. Unlike \textsc{mon}et, \gls{GENESIS} clearly differentiates individual objects in the first example. In the second example, \gls{GENESIS} captures the fine-grained pattern of the wall in the background better than \textsc{mon}et.}
    \label{fig:decomposition_gqn}
\end{figure}

Like \textsc{mon}et and \textsc{iodine}, which were designed for unsupervised scene decomposition, \gls{GENESIS} is also able to segment scenes into meaningful components.
\Cref{fig:decomposition_gqn} compares the decomposition of two images from the GQN dataset with \gls{GENESIS} and \textsc{mon}et.
Both models follow a similar decomposition strategy, but \textsc{mon}et fails to disambiguate one foreground object in the first example and does not reconstruct the background in as much detail in the second example.
In \Cref{app:scene_decomposition}, \Cref{fig:inference_gqn_same_colour} illustrates the ability of both methods to disambiguate objects of the same colour and \Cref{fig:inference_gqn_gs} shows scene decomposition with \gls{GENESIS}\textsc{-s}.

Following \citet{greff2019multi}, we quantify segmentation performance with the Adjusted Rand Index (ARI) of pixels overlapping with ground truth foreground objects.
We computed the ARI on 300 random images from the ShapeStacks test set for five models trained with different random seeds.
\mbox{\gls{GENESIS}} achieves an ARI of $0.73\pm0.03$ which is better than $0.63\pm0.07$ for \textsc{mon}et.
This metric, however, does not penalise objects being over-segmented, which can give a misleading impression with regards to segmentation quality.
This is illustrated in \Cref{fig:decomposition_shapestacks_h4}, \Cref{app:scene_decomposition}.

Inspired by \citet{arbelaez2010contour}, we thus propose to use the \emph{segmentation covering} (SC) of the ground truth foreground objects by the predicted masks.
This involves taking a weighted mean over mask pairs, putting a potentially undesirable emphasis on larger objects.
We therefore also consider taking an unweighted mean (mSC).
For the same 300 images from the ShapeStacks test set and five different random seeds, \gls{GENESIS} (SC: $0.64\pm0.08$, mSC: $0.60\pm0.09$) again outperforms \textsc{mon}et (SC: $0.52\pm0.09$, mSC: $0.49\pm0.09$).
More details are provided in \Cref{app:segmentation_covering}.

\subsection{Evaluation of Unsupervised Representation Utility}
\label{sec:semi}

Using a subset of the available labelled training images from ShapeStacks, we train a set of classifiers on the representations learned by \gls{GENESIS} and several baselines to evaluate how well these representations capture the ground truth scene state.
In particular, we consider three tasks: (1) Is a tower stable or not? (2) What is the tower's height in terms of the number of blocks? (3) What is the camera viewpoint (out of 16 possibilities)?
Tower stability is a particularly interesting property as it depends on in fine-grained object information and the relative positioning of objects.
We selected the third task as learning scene representations from different views has previously been prominently explored in \citet{eslami2018neural}.
We compare \gls{GENESIS} and \gls{GENESIS}\textsc{-s} against three baselines: \textsc{mon}et, a \gls{VAE} with a spatial broadcast decoder (\textsc{bd-vae}) and a \gls{VAE} with a deconvolutional decoder (\textsc{dc-vae}).
The results are summarised in \Cref{tab:semisupervised}.
The architectural details of the baselines are described in \Cref{app:monet_baseline} and \Cref{app:vae_baseline}.
The implementation details of the classifiers are provided in \Cref{app:classifiers}.

\clearpage

Both \gls{GENESIS} and \gls{GENESIS}\textsc{-s} perform better than than the baselines at predicting tower stability and their accuracies on predicting the height of the towers is only outperformed by \textsc{mon}et.
We conjecture that \textsc{mon}et benefits here by its deterministic segmentation network.
Overall, this corroborates the intuition that object-centric representations are indeed beneficial for these tasks which focus on the foreground objects.
We observe that the \textsc{bd-vae} does better than the \textsc{dc-vae} on all three tasks, reflecting the motivation behind its design which is aimed at better disentangling the underlying factors of variation in the data \citep{watters2019spatial}.
All models achieve a high accuracy at predicting the camera view.
Finally, we note that none of models reach the stability prediction accuracies reported in \citet{groth2018shapestacks} which were obtained with an Inception-v4 classifier \citep{szegedy2017inception}.
This is not surprising considering that only a subset the training images is used for training the classifiers without data augmentation and at a reduced resolution.

\begin{table}[h!]
    \centering
    \caption{Classification accuracy in \% on the test sets of the ShapeStacks tasks.}
    \label{tab:semisupervised}
    \newcolumntype{Y}{>{\centering\arraybackslash}X}
    \begin{tabularx}{0.85\textwidth}{l c*{6}{Y}}
        \toprule
        Task & \gls{GENESIS} & \mbox{\gls{GENESIS}\textsc{-s}} & \textsc{mon}et & \textsc{bd-vae} & \textsc{dc-vae} & Random \\
        \cmidrule(r){1-1} \cmidrule(rl){2-4} \cmidrule(rl){5-6} \cmidrule(r){7-7}
        Stability & \textbf{64.0} & 63.2          & 59.6          & 60.1          & 59.0 & 50.0 \\
        Height    & 80.3          & 80.8          & \textbf{88.4} & 78.6          & 67.5 & 22.8 \\
        View      & 99.3          & \textbf{99.7} & 99.5          & \textbf{99.7} & 99.1 & 6.25 \\
        \bottomrule
    \end{tabularx}
\end{table}

\subsection{Quantifying Sample Quality}
\label{sec:vae}

In order to quantify the quality of generated scenes, \Cref{tab:fid} summarises the Fr\'{e}chet Inception Distances (FIDs) \citep{heusel2017gans} between 10,000 images generated by \gls{GENESIS} as well several baselines and 10,000 images from the Multi-dSprites and the GQN test sets, respectively.
The two \gls{GENESIS} variants achieve the best FID on both datasets.
While \gls{GENESIS}\textsc{-s} performs better than \gls{GENESIS} on GQN, \Cref{fig:generation_gqn_gs6} and \Cref{fig:generation_gqn_gs_4x8} in \Cref{app:generation} show that individual scene components are less interpretable and that intricate background patterns are generated at the expense of sensible foreground objects.
It is not surprising that the FIDs for \textsc{mon}et are relatively large given that it was not designed for generating scenes.
Interestingly, the \textsc{dc-vae} achieves a smaller FID on GQN than the \textsc{bd-vae}.
This is surprising given that the \textsc{bd-vae} representations are more useful for the ShapeStacks classification tasks.
Given that the GQN dataset and ShapeStacks are somewhat similar in structure and appearance, this indicates that while FID correlates with perceptual similarity, it does not necessarily correlate with the general utility of the learned representations for downstream tasks.
We include scenes sampled from the \textsc{bd-vae} and the \textsc{dc-vae} in \Cref{fig:generation_gqn_vae}, \Cref{app:generation}, where we observe that the \textsc{dc-vae} models the background fairly well while foreground objects are blurry.

\begin{table}[h!]
    \centering
    \caption{Fr\'{e}chet Inception Distances for \gls{GENESIS} and baselines on GQN.}
    \label{tab:fid}
    \newcolumntype{Y}{>{\centering\arraybackslash}X}
    \begin{tabularx}{0.8\textwidth}{l c*{4}{Y}}
        \toprule
        Dataset & \gls{GENESIS} & \mbox{\gls{GENESIS}\textsc{-s}} & \textsc{mon}et & \textsc{bd-vae} & \textsc{dc-vae} \\
        \cmidrule(r){1-1} \cmidrule(lr){2-4} \cmidrule(r){5-6}
        Multi-dSprites & \textbf{24.9} & 28.2          & 92.7  & 89.8  & 100.5 \\
        GQN            & 80.5          & \textbf{70.2} & 176.4 & 145.5 & 82.5 \\
        \bottomrule
    \end{tabularx}
\end{table}

\section{Conclusions}
\label{sec:conclusions}

In this work, we propose a novel object-centric latent variable model of  scenes called \gls{GENESIS}.
We show that \gls{GENESIS} is, to the best of our knowledge, the first unsupervised model to both decompose rendered 3D scenes into semantically meaningful constituent parts, while at the same time being able to generate coherent scenes in a component-wise fashion.
This is achieved by capturing relationships between scene components with an autoregressive prior that is learned alongside a computationally efficient sequential inference network, setting \gls{GENESIS} apart from prior art.
Regarding future work, an interesting challenge is to scale \gls{GENESIS} to more complex datasets and to employ the model in robotics or reinforcement learning applications.
To this end, it will be necessary to improve reconstruction and sample quality, reduce computational cost, and to scale the model to higher resolution images.
Another potentially promising research direction is to adapt the formulation to only model parts of the scene that are relevant for a certain task.

\clearpage

\subsubsection*{Acknowledgments}
This research was supported by an EPSRC Programme Grant (EP/M019918/1), an EPSRC DTA studentship, and a Google studentship.
The authors would like to acknowledge the use of the University of Oxford Advanced Research Computing (ARC) facility in carrying out this work, \url{http://dx.doi.org/10.5281/zenodo.22558}, and the use of Hartree Centre resources.
The authors would like to thank Yizhe Wu for his help with re-implementing \textsc{mon}et, Oliver Groth for his support with the GQN and ShapeStacks datasets, and Rob Weston for proof reading the paper.

\bibliography{references}  %
\bibliographystyle{iclr2020_conference}

\clearpage
\appendix

\section{Datasets}
\label{app:datasets}

\textbf{Multi-dSprites \citep{burgess2019monet}}\ \
Images contain between one and four randomly selected ``sprites'' from \citet{matthey2017dsprites}, available at \url{https://github.com/deepmind/dsprites-dataset}.
For each object and the background, we randomly select one of five different, equally spread values for each of the three colour channels and generate 70,000 images.
We set aside 10,000 for validation and testing each.
The script for generating this data will be released with the rest of our code.

\textbf{GQN \citep{eslami2018neural}}\ \
The ``rooms-ring-camera'' dataset includes simulated 3D scenes of a square room with different floor and wall textures, containing one to three objects of various shapes and sizes.
It can be downloaded from \url{https://github.com/deepmind/gqn-datasets}.

\textbf{ShapeStacks \citep{groth2018shapestacks}}\ \
Images show simulated block towers of different heights (two to six blocks).
Individual blocks can have different shapes, sizes, and colours.
Scenes have annotations for: stability of the tower (binary), number of blocks (two to six), properties of individual blocks, locations in the tower of centre-of-mass violations and planar surface violations, wall and floor textures (five each), light presets (five), and camera view points (sixteen).
More details about the dataset and download links can be found at \url{https://shapestacks.robots.ox.ac.uk/}.

\section{Implementation Details}
\label{app:implementation}

\subsection{Genesis Architecture}
\label{app:genesis_architecture}

We use the architecture from \citet{berg2018sylvester} to encode and decode $\bz^m_{k}$ with the only modification of applying batch normalisation \citep{ioffe2015batch} before the \textsc{glu} non-linearities \citep{dauphin2017language}.
The convolutional layers in the encoder and decoder have five layers with size-5 kernels, strides of [1, 2, 1, 2, 1], and filter sizes of [32, 32, 64, 64, 64] and [64, 32, 32, 32, 32], respectively.
Fully-connected layers are used at the lowest resolution.

The encoded image is passed to a \gls{LSTM} cell \citep{hochreiter1997long} followed by a linear layer to compute the mask latents $\bz^m_{k}$ of size 64.
The \gls{LSTM} state size is twice the latent size.
Importantly, unlike the analogous counterpart in \textsc{mon}et, the decoding of $\bz^m_{k}$ is performed in parallel.
The autoregressive prior $\p{\bz^m_k}{\bz^m_{1:k-1}}{\theta}$ is implemented as an \gls{LSTM} with 256 units.
The conditional distribution $\p{\bz^c_k}{\bz^m_k}{\theta}$ is parameterised by a \gls{MLP} with two hidden layers, 256 units per layer, and \textsc{elu}s \citep{clevert2015fast}.
We use the same component \gls{VAE} featuring a spatial broadcast decoder as \textsc{mon}et to encode and decode $z^c_k$, but we replace \textsc{relu}s \citep{glorot2011deep} with \textsc{elu}s.

\begin{figure}[h]
    \centering
    \includegraphics[trim=0 0 0 0, clip, width=0.5\textwidth]{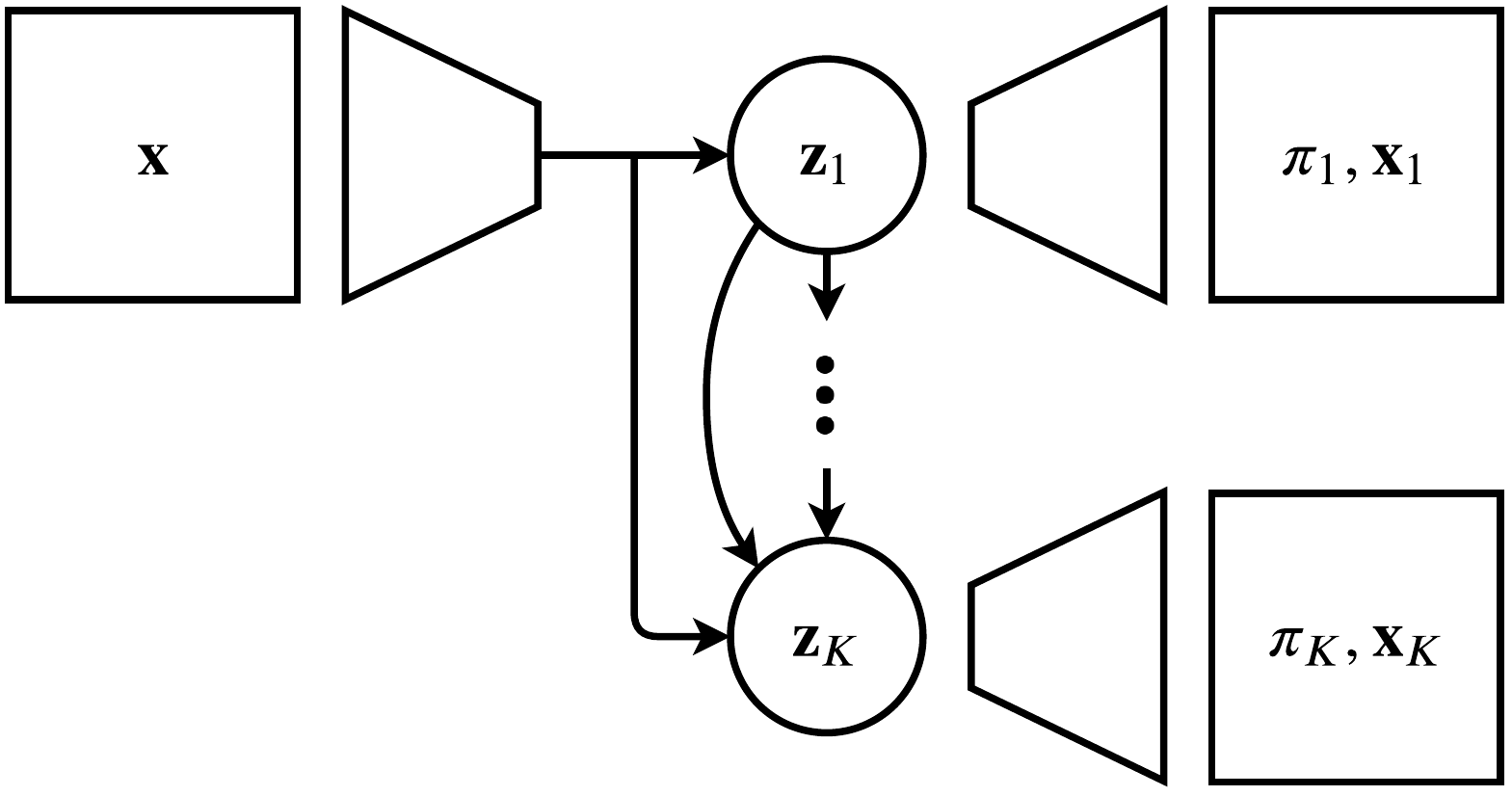}
    \caption{\gls{GENESIS}-\textsc{s} overview. Given an image $\bx$, an encoder and an \gls{RNN} compute latent variables $\bz_k$. These are decoded to directly obtain the mixing probabilities $\mathbf{\pi}_k$ and the scene components $\bx_k$.}
    \label{fig:arch_s}
\end{figure}

For \gls{GENESIS}\textsc{-s}, as illustrated in \Cref{fig:arch_s}, the encoder of $\bz_{k}$ is the same as for $\bz^m_{k}$ above and the decoder from \citet{berg2018sylvester} is again used to compute the mixing probabilities.
However, \gls{GENESIS}\textsc{-s} also has a second decoder with spatial broadcasting to obtain the scene components $\bx_{k}$ from $\bz_{k}$.
We found the use of two different decoders to be important for \gls{GENESIS}\textsc{-s} in order for the model to decompose the input.

\subsection{MONet Baselines}
\label{app:monet_baseline}

We followed the provided architectural details described in \citet{burgess2019monet}.
Regarding unspecified details, we employ an attention network with [32, 32, 64, 64, 64] filters in the encoder and the reverse in the decoder.
Furthermore, we normalise the mask prior with a $\operatorname{softmax}$ function to compute the \textsc{kl}-divergence between mask posterior and prior distributions.

\subsection{VAE Baselines}
\label{app:vae_baseline}

Both the \textsc{bd-vae} and the \textsc{dc-vae} have a latent dimensionality of 64 and the same encoder as in \citet{berg2018sylvester}.
The \textsc{dc-vae} also uses the decoder from  \citet{berg2018sylvester}.
The \textsc{bd-vae} has the same spatial broadcast decoder with \textsc{elu}s as \gls{GENESIS}, but with twice the number of filters to enable a better comparison.

\subsection{Optimisation}

The scalar standard deviation of the Gaussian image likelihood components is set to $\sigma_x = 0.7$.
We use \gls{GECO} \citep{rezende2018taming} to balance the reconstruction and KL divergence terms in the loss function.
The goal for the reconstruction error is set to $0.5655$, multiplied by the image dimensions and number of colour channels.
We deliberately choose a comparatively weak reconstruction constraint for the \gls{GECO} objective to emphasise \textsc{kl} minimisation and sample quality.
For the remainining \gls{GECO} hyperparameters, the default value of $\alpha = 0.99$ is used and the step size for updating $\beta$ is set to $10^{-5}$.
We increase the step size to $10^{-4}$ when the reconstruction constraint is satisfied to accelerate optimisation as $\beta$ tended to undershoot at the beginning of training.

All models are trained for $5*10^{5}$ iterations with a batch size of 32 using the \gls{ADAM} optimiser \citep{kingma2014adam} and a learning rate of $10^{-4}$.
With these settings, training \gls{GENESIS} takes about two days on a single GPU.
However, we expect performance to improve with further training.
This particularly extends to training \gls{GENESIS} on ShapeStacks where $5*10^{5}$ training iterations are not enough to achieve good sample quality.

\subsection{ShapeStacks Classifiers}
\label{app:classifiers}

Multilayer perceptrons (\textsc{mlp}s) with one hidden layer, 512 units, and \textsc{elu} activations are used for classification.
The classifiers are trained for 100 epochs on 50,000 labelled examples with a batch size of 128 using a cross-entropy loss, the \gls{ADAM} optimiser, and a learning rate of $10^{-4}$.
As inputs to the classifiers, we concatenate $\bz^m_k$ and $\bz^c_k$ for \gls{GENESIS}, $\bz_k$ for \gls{GENESIS}\textsc{-s}, and the component \gls{VAE} latents for the two \textsc{mon}et variants.

\section{Segmentation Covering}
\label{app:segmentation_covering}

\begin{figure}[h]
    \centering
    \includegraphics[width=1.0\textwidth]{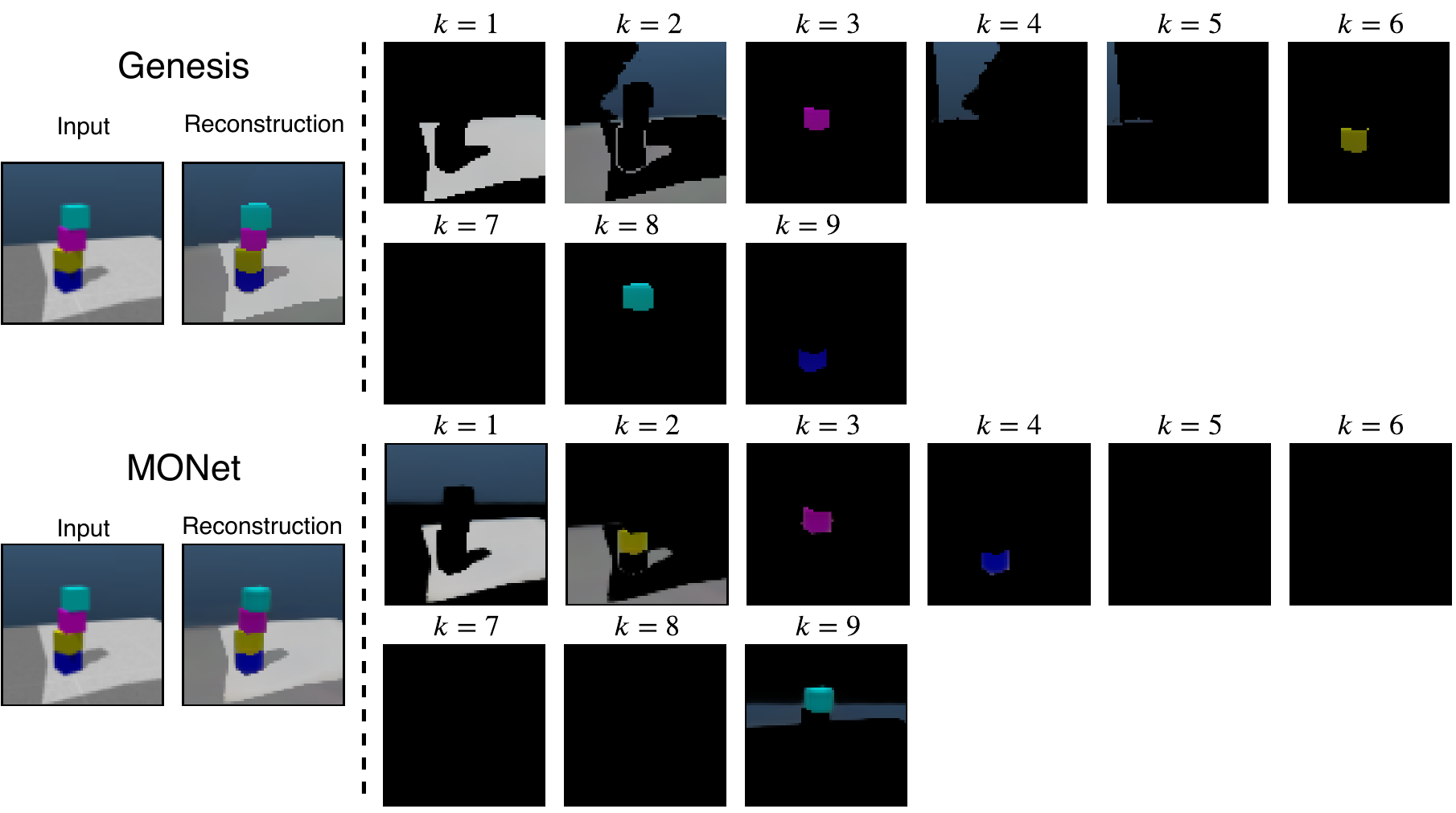}
    \caption{In this example, \gls{GENESIS} (ARI: 0.83, SC: 0.83, mSC: 0.83) segments the four foreground objects properly.
    \textsc{mon}et (ARI: 0.89, SC: 0.47, mSC: 0.50), however, merges foreground objects and background again in steps $k=2$ and $k=9$.
    Despite the inferior decomposition, the ARI for \textsc{mon}et is higher than for \mbox{\gls{GENESIS}}. This is possible as the ARI does not penalise the over-segmentation of the foreground objects, highlighting its limitations for evaluating unsupervised instance segmentation. The segmentation covering, however, reflects the quality of the segmentatioin masks properly.}
    \label{fig:decomposition_shapestacks_h4}
\end{figure}

Following \citet{arbelaez2010contour}, the \emph{segmentation covering} (SC) is based on the \gls{IOU} between pairs of segmentation masks from two sets $S$ and $S'$.
In this work, we consider $S$ to be the segmentation masks of the ground truth foreground objects and $S'$ to be the predicted segmentation masks.
The covering of $S$ by $S'$ is defined as:
\begin{equation}
    C(S'\rightarrow S) = \frac{1}{\sum_{R \in S} \abs{R}} \sum_{R \in S} \abs{R} \max_{R' \in S'} \operatorname{\textsc{iou}}(R, R'),
\end{equation}
where $\abs{R}$ denotes the number of pixels belonging to mask $R$.
Note that this formulation is slightly more general than the one in \citet{arbelaez2010contour} which assumes that masks in $S$ are non-overlapping and cover the entire image.
The above takes a weighted mean over \gls{IOU} values, proportional to the number of pixels of the masks being covered.
To give equal importance to masks of different sizes, we also consider taking an unweighted mean (mSC):
\begin{equation}
    C_m(S'\rightarrow S) = \frac{1}{\abs{S}} \sum_{R \in S}  \max_{R' \in S'} \operatorname{\textsc{iou}}(R, R'),
\end{equation}
where $\abs{S}$ denotes the number of non-empty masks in $S$. Importantly and unlike the ARI, both segmentation covering variations penalise the over-segmentation of ground truth objects as this decreases the \gls{IOU} for a pair of masks.
This is illustrated in \Cref{fig:decomposition_shapestacks_h4}, \Cref{app:scene_decomposition}.

\clearpage

\section{Component-Wise Scene Generation - GQN}
\label{app:generation}

\begin{figure}[h]
    \centering
    \includegraphics[width=1.0\textwidth]{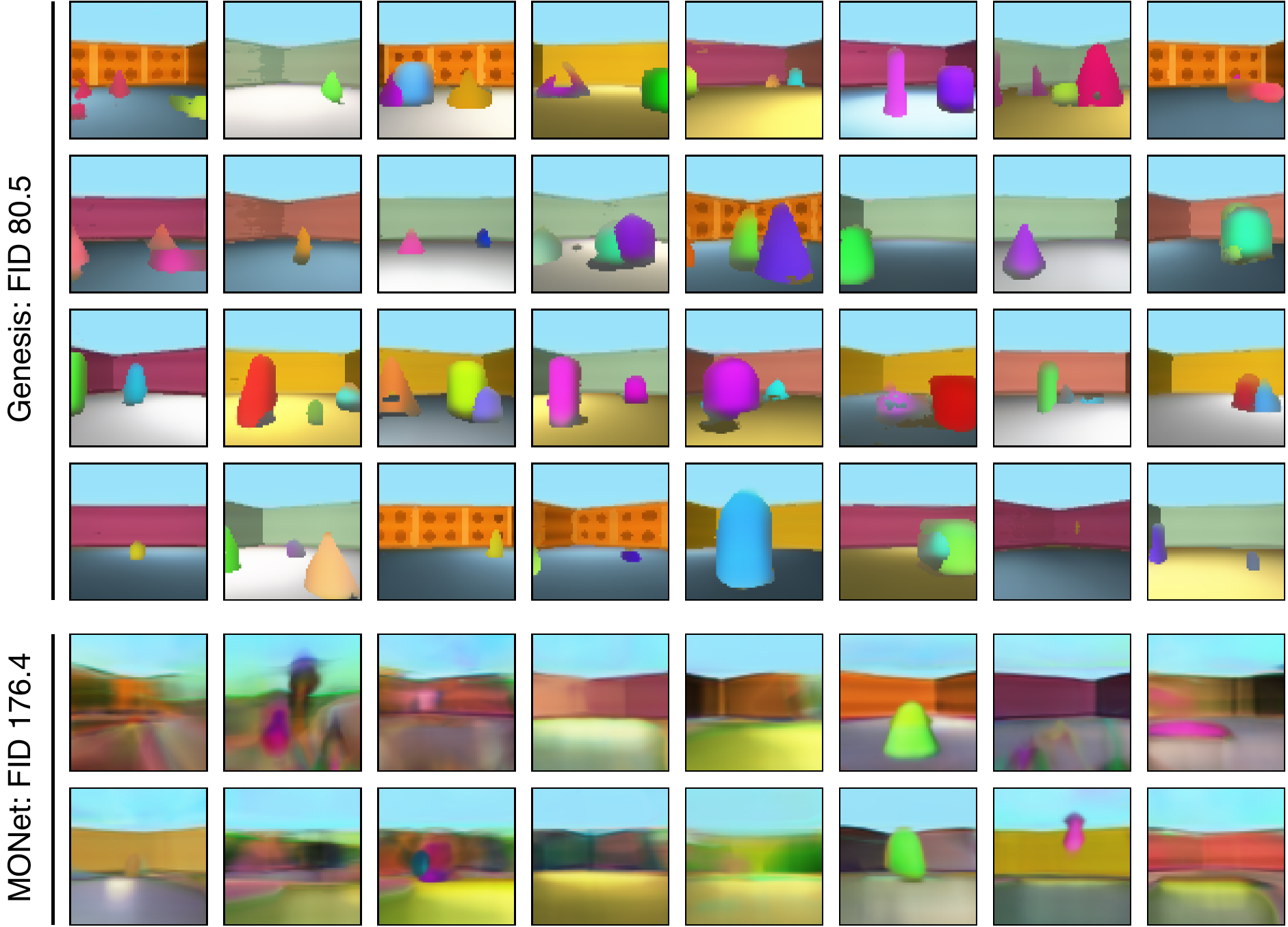}
    \caption{Randomly selected scenes generated by \gls{GENESIS} and \textsc{mon}et after training on the GQN dataset.
    Images sampled from \gls{GENESIS} contain clearly distinguishable foreground objects and backgrounds. Samples from \textsc{mon}et, however, are mostly incoherent.}
    \label{fig:generation_gqn_g32m16}
\end{figure}

\begin{figure}[h]
    \centering
    \includegraphics[width=1.0\textwidth]{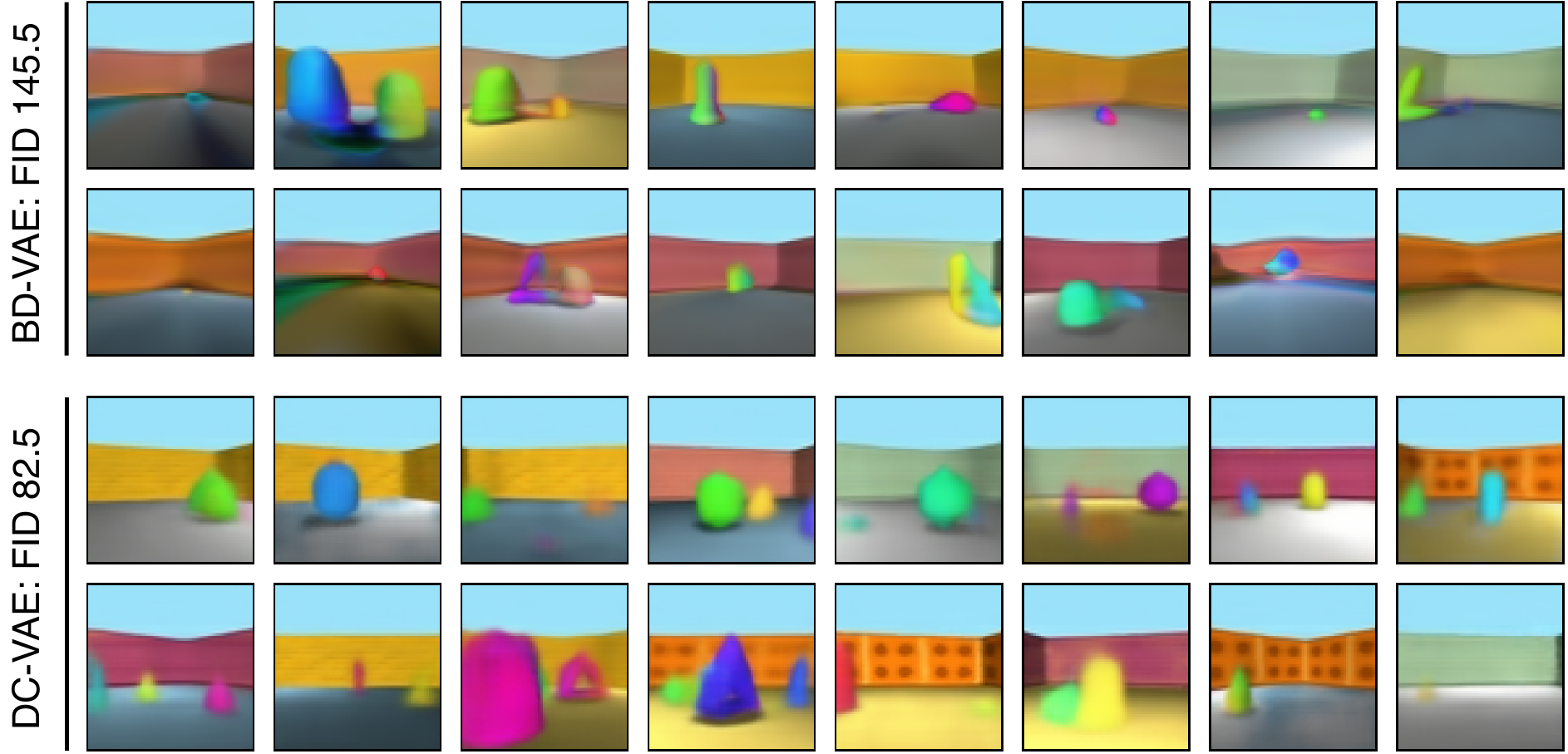}
    \caption{Randomly selected scenes generated by the BD-VAE and the DC-VAE after training on
the GQN dataset; shown for comparison. The DC-VAE generates decent scene backgrounds but foreground objects are blurry.}
    \label{fig:generation_gqn_vae}
\end{figure}

\begin{figure}[h]
    \centering
    \includegraphics[width=1.0\textwidth]{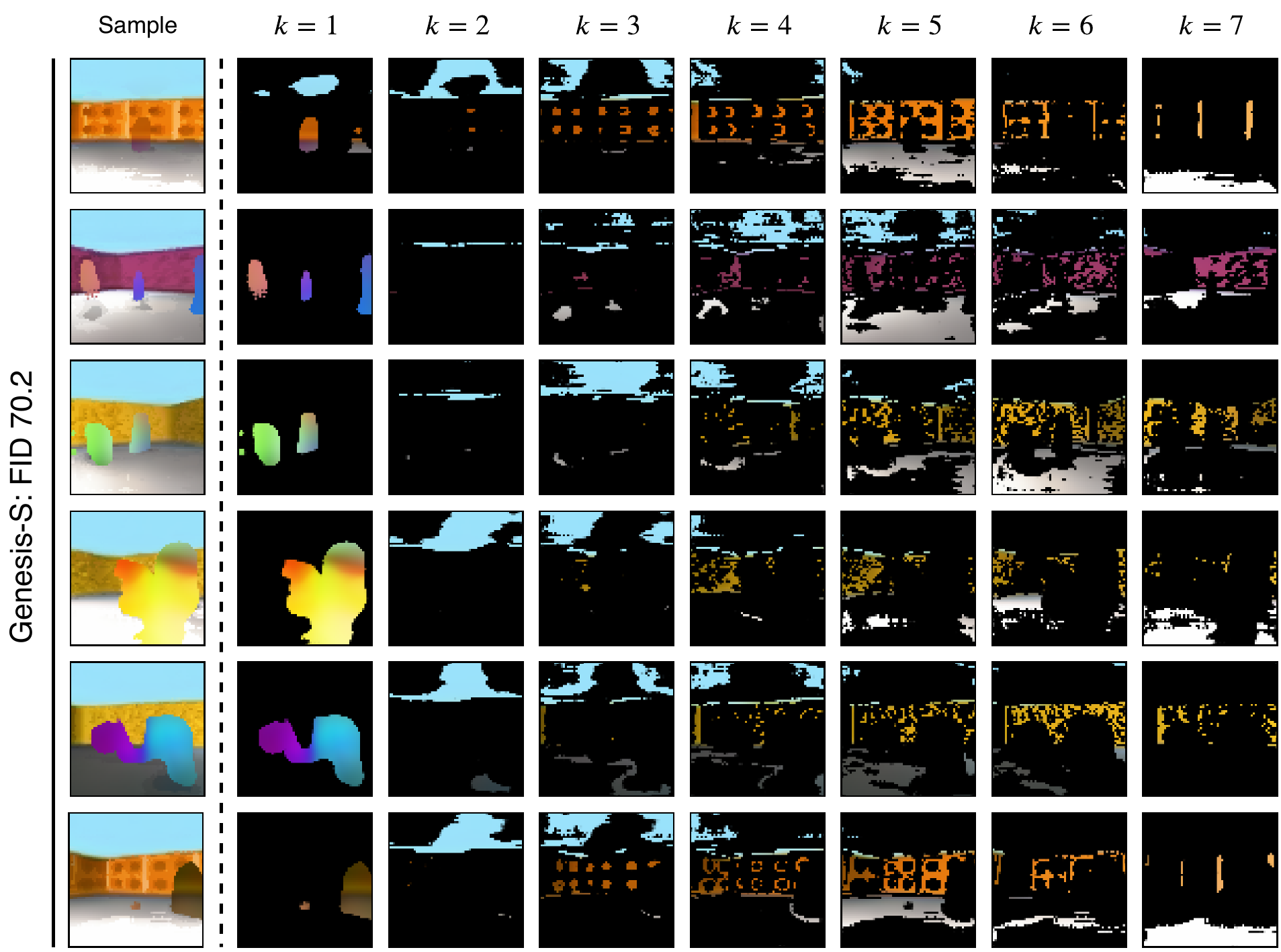}
    \caption{Component-by-component scene generation with \gls{GENESIS}\textsc{-s} after training on the GQN dataset. While \gls{GENESIS}\textsc{-s} nominally achieves the best FID in \Cref{tab:fid}, this appears to be due to the generation of high fidelity background patterns rather than appropriate foreground objects. Furthermore, unlike the components generated by \gls{GENESIS} at every step in \Cref{fig:generation_gqn}, the components generated by \gls{GENESIS}\textsc{-s} are not very interpretable.}
    \label{fig:generation_gqn_gs6}
\end{figure}

\begin{figure}[h]
    \centering
    \includegraphics[width=1.0\textwidth]{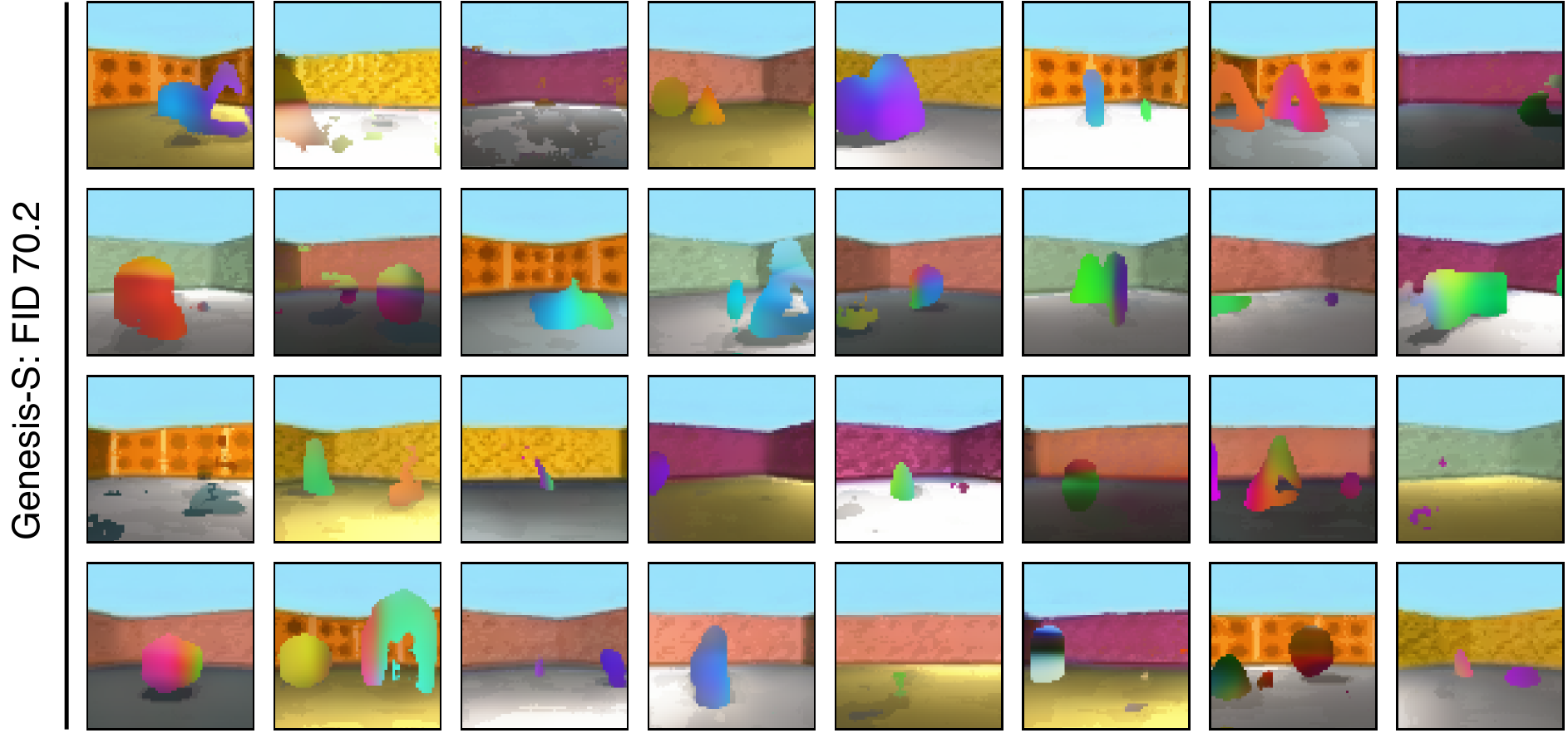}
    \caption{Randomly selected scenes generated by \gls{GENESIS}\textsc{-s} after training on the GQN dataset.}
    \label{fig:generation_gqn_gs_4x8}
\end{figure}

\clearpage

\section{Inference of Scene Components}
\label{app:scene_decomposition}

\begin{figure}[h]
    \centering
    \includegraphics[width=1.0\textwidth]{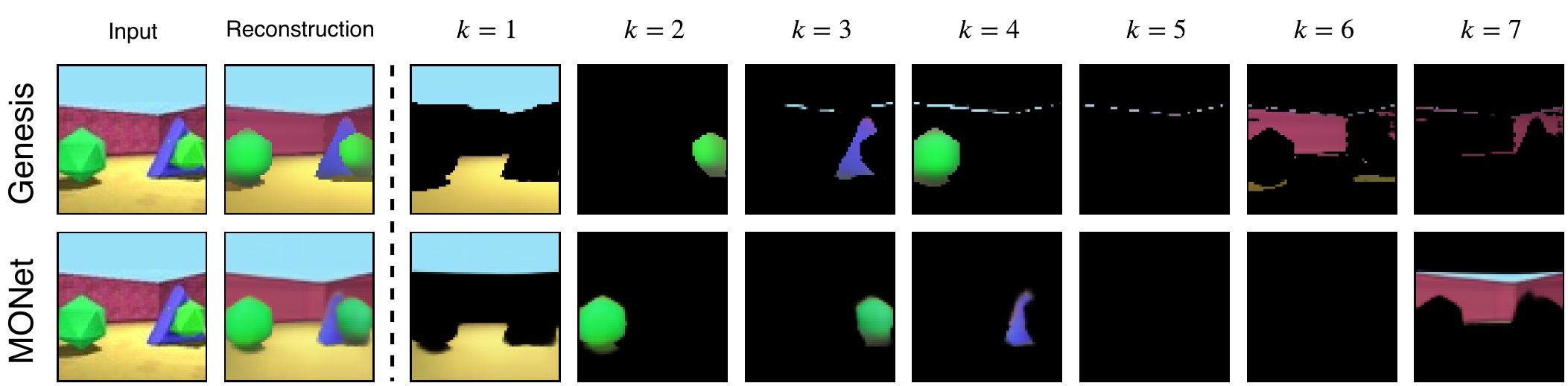}
    \caption{Step-by-step decomposition of a scene from GQN with \gls{GENESIS} and \textsc{mon}et. Two objects with the same shape and colour are successfully identified by both models. While colour and texture are useful cues for decomposition, this example shows that both models perform something more useful than merely identifying regions of similar colour.}
    \label{fig:inference_gqn_same_colour}
\end{figure}

\begin{figure}[h]
    \centering
    \includegraphics[width=1.0\textwidth]{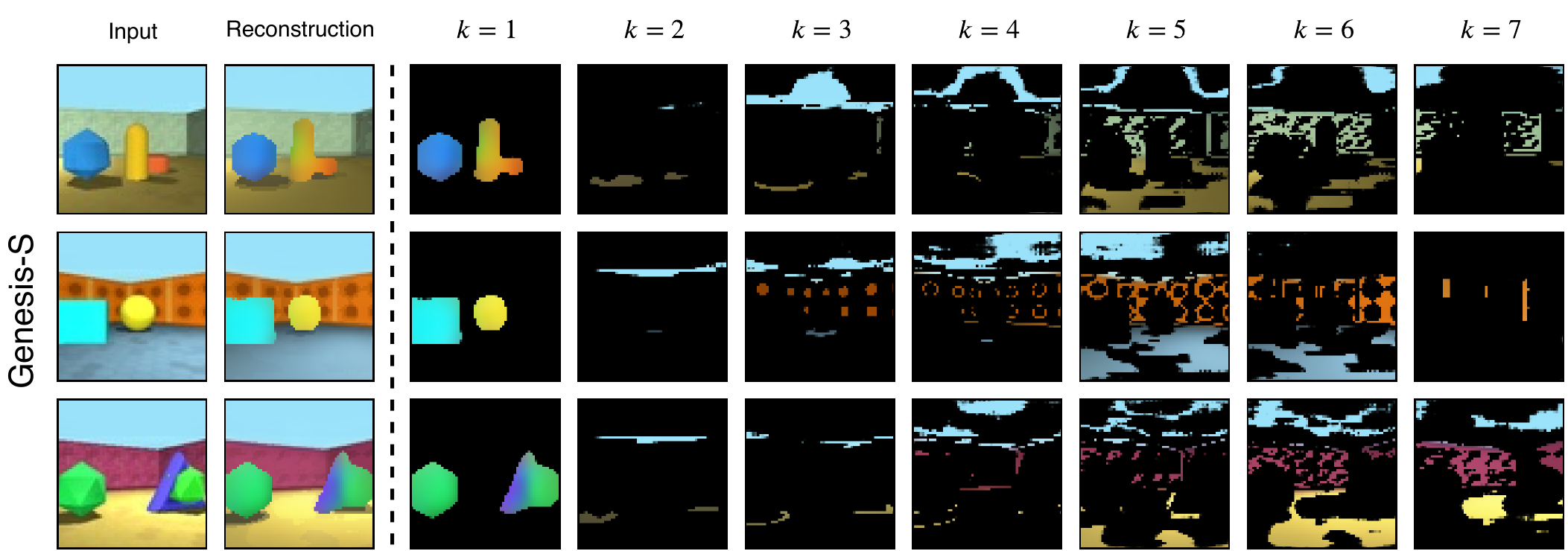}
    \caption{Step-by-step decomposition of the same scenes as in \Cref{fig:decomposition_gqn} and \Cref{fig:inference_gqn_same_colour} with \mbox{\gls{GENESIS}\textsc{-s}}. While the foreground objects are distinguished from the background, they are explained together in the first step. Subsequent steps reconstruct the background in a haphazard fashion.}
    \label{fig:inference_gqn_gs}
\end{figure}

\begin{figure}[h]
    \centering
    \includegraphics[width=1.0\textwidth]{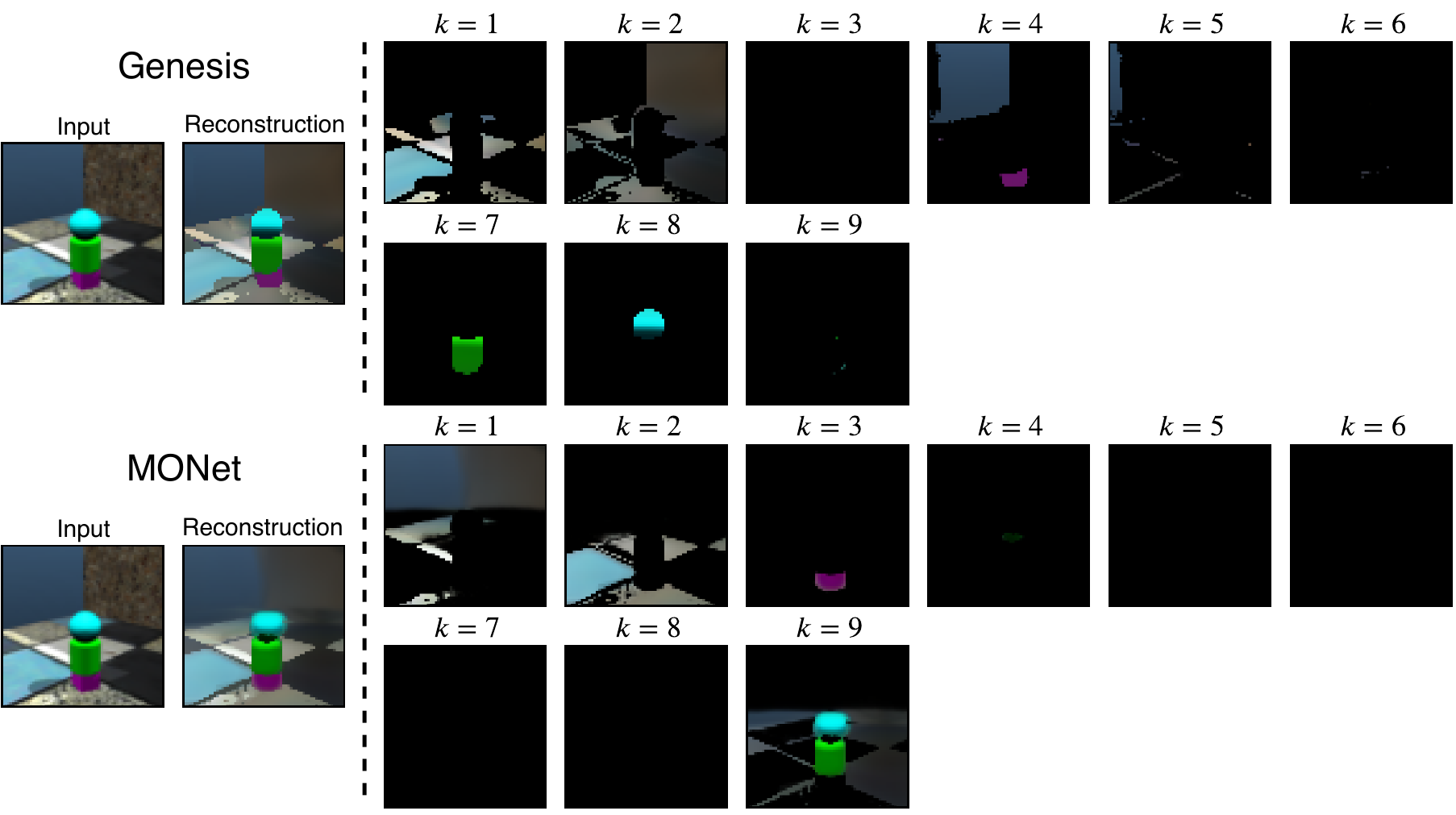}
    \caption{A ShapeStacks tower is decomposed by \gls{GENESIS} and \textsc{mon}et.
    Compared to the GQN dataset, both methods struggle to segment the foreground objects properly. \gls{GENESIS} captures the purple shape and parts of the background wall in step $k=4$. \textsc{mon}et explains the green shape, the cyan shape, and parts of floor in step $k=9$.
    This is reflected in the foreground ARI and segmentation covering for \gls{GENESIS} (ARI: 0.82, SC: 0.68, mSC: 0.58) and \textsc{mon}et (ARI: 0.39, SC: 0.26, mSC: 0.35); the latter being lower as the green and cyan shapes are not separated.}
    \label{fig:decomposition_shapestacks_h3}
\end{figure}

\end{document}